\documentclass{article}

\PassOptionsToPackage{numbers, sort&compress}{natbib}
\usepackage[dblblindworkshop, final]{neurips_2026}
\workshoptitle{AXIOM: Foundations of Efficient Deep Learning}

\usepackage[table,svgnames,x11names]{xcolor}
\definecolor{citecolor}{HTML}{0071bc}
\usepackage[pagebackref=false,breaklinks=true,letterpaper=true,colorlinks,citecolor=citecolor,bookmarks=false]{hyperref}
\usepackage[utf8]{inputenc} % allow utf-8 input
\usepackage[T1]{fontenc}    % use 8-bit T1 fonts
\usepackage{url}            % simple URL typesetting
\usepackage{booktabs}       % professional-quality tables
\usepackage{amsfonts}       % blackboard math symbols
\usepackage{nicefrac}       % compact symbols for 1/2, etc.
\usepackage{microtype}      % microtypography
\usepackage{xcolor}         % colors
\usepackage{natbib}
\usepackage{amsmath}
\usepackage{multirow}
\usepackage{graphicx}
\usepackage{algorithmic}
\usepackage{colortbl}
\title{LeanTTA: A Backpropagation-Free, Stateless Approach to On-Device Test-Time Adaptation}

\newcommand{\repourl}{https://github.com/cam-mobsys/LeanTTA}

\author{Cynthia Dong\textsuperscript{1} \quad
  Young D. Kwon\textsuperscript{2,4} \quad
  Hong Jia\textsuperscript{3} \quad
  George Rizos\textsuperscript{2} \quad
  Cecilia Mascolo\textsuperscript{2} \\
  \textsuperscript{1}University of Washington \quad \textsuperscript{2}University of Cambridge \quad \textsuperscript{3}University of Auckland \\
  \textsuperscript{4}Samsung AI Center-Cambridge, UK \\
  \texttt{cyndo@cs.washington.edu} \quad \texttt{yd.kwon@samsung.com}
}

\newcommand{\new}[1]{\textcolor{black}{#1}}

\begin{document}

\maketitle

\begin{abstract}
  
For small, compressed models constrained by edge hardware and limited data storage or availability, test-time adaptation (TTA) can restore robustness against common corruptions that induce covariate shift. However, many TTA methods rely on backpropagation and large batch sizes, which we show fail under constrained data and hardware. In this work, we present \textbf{LeanTTA}, a combination of memory and compute-efficient tactics which allow us to deploy TTA on-device, even under quantized paradigms. LeanTTA consists of two components: (1) stateless, backpropagation-free statistics adjustment using an efficient calculation of the Mahalanobis distance and (2) partial adaptation through layerwise updates. LeanTTA reaches up to a 16\% reduction in corruption error on CIFAR10 with minimal additional compute time and memory required over normal inference, demonstrating the potential of backpropagation-free methods for efficient on-device TTA.

Code: \href{\repourl}{\repourl}.

\end{abstract}

\section{Introduction}
Test-Time Adaptation (TTA) promises robust model deployment under distribution shifts, without access to data labels. For small models compressed to fit on edge hardware, TTA on-device can provide extra adaptability lost to quantization or compression---important, given that edge deployments encounter in-the-wild data which may be not only missing from the training set, but also highly limited in quantity, such as one recording of a rare bird under unusual conditions, or a single photo of a rare skin condition in a remote clinic~\citep{dai19_skin,stowell18_birds,koh20_wilds}. However, mainstream approaches rely heavily on backpropagation and large batch sizes~\citep{wang21_tent}, which become untenable within edge limitations, given memory constraints, the instability of quantized fine-tuning, and low data availability. We find in this work that small batch sizes can in fact collapse model accuracy to random guessing with most backpropagation-driven TTA methods, illustrating the incompatibility of most TTA methods and on-device deployment.

To address this gap, we introduce \textbf{LeanTTA}, a combination of memory- and compute-efficient tactics which allow us to deploy TTA on-device, even under low-data and quantized paradigms. We first present stateless, backpropagation-free statistics adjustment via Mahalanobis distance, then demonstrate the accuracy and efficiency benefits of selective layerwise adaptation, which allow for fusion of unadapted layers for further computational gains. We evaluate on a common edge architecture (ResNet18~\citep{he16_resnet}, others in Appendix) and multiple datasets (ImageNet-C, CIFAR10-C~\citep{hendrycks19_dsets}, and Warblr~\citep{stowell18_birds}), benchmarking against SOTA methods on a resource-limited Raspberry Pi Zero 2W. Evaluated across corruption benchmarks, LeanTTA achieved up to a 16\% reduction in corruption error ($\frac{Adaptation - Corruption}{Baseline - Corruption}$, calculated on CIFAR-10 with a baseline of 93\%) \textit{across both gradual and abrupt shifts} while operating within an 11.2~MB peak memory footprint on ResNet18, delivering an order-of-magnitude reduction in memory consumption over similarly-performing TTA methods.
\section{Background and Related Works}
\label{sec:bgd}
TTA aims to reduce the effect of test-time covariate shift through fine-tuning or direct manipulation of the source statistics frozen in batch normalization layers at test time~\citep{koh20_wilds}. Normalization layers ensure that intermediate activations arrive at subsequent layers with a standardized mean and variance~\citep{ioffe15_bn, santurkar18_bn}; when deployment-time data arrives with shifted statistics, frozen training statistics cannot standardize activations into the operating region expected by other layers~\citep{li16_adabn}.

The early work Tent~\citep{wang21_tent} and many variants~\citep{wang22_cotta, niu23_sar, niu22_eata, lee24_deyo} rely on entropy-supervised backpropagation, where batch normalization statistics are \textit{continuously} updated using gradient descent, supervised by entropy (and thereby confidence) as a proxy for accuracy drops due to covariate shift. Methods like~\citep{niu22_eata, lee24_deyo, niu23_sar} iterate on entropy-backpropagation methods through selective gradient updates, while methods like ~\citep{wang22_cotta, yuan23_rotta, gong23_sotta} aim to decrease model collapse and error accumulation due to continual adaptation. For example, MECTA~\citep{hong23_mecta} aims to reduce memory through pruning and selective activation caching, and TinyTTA~\citep{jia2024tinytta} through early-exit
ensembles on microcontrollers. These works require backwards passes and gradient calculations---untenable with limited compute.

Forward-only adaptation as proposed in~\citep{li16_adabn, su24_tema,niu24_foa,ma26_pea} improves memory- and compute efficiency. TEMA~\citep{su24_tema} estimates statistics over previous test batches, FOA~\citep{niu24_foa} relies on a particular architecture, and PEA~\citep{ma26_pea} lifts this restriction yet relies on an exponential moving average of test-time statistics.

LeanTTA builds on the efficiency of forward-only methods and adds several features. To mitigate degradation from continual adaptation, we demonstrate that adaptation relative to \textit{original} source statistics \textit{per-instance} is sufficient to improve accuracy. LeanTTA also leverages for more compute efficiency the finding that shallow-layer adaptation matches or exceeds full-model adaptation, and proposes fusion of non-adaptive layers for further memory and compute savings (see extended discussion in Appendix~\ref{app:related}).

\section{The Method}

\paragraph{\textbf{Problem Description:}} A model $f_{\theta}(\textbf{x})$ is trained over a source dataset with inputs and labels $x_s, y_s$, sampled from the joint probability distribution $P_{source}(x_s, y_s)$. At test-time, the model $f_{\theta}(\textbf{x})$ with parameters $\theta$ encounters inputs $x_t$ with unknown labels $y_t$, where $x_t, y_t$ are distributed over $P_{target}(x_t, y_t)$, and $P_{target}(x_t, y_t)$ may have diverged from $P_{source}(x_s, y_s)$. The model $f_{\theta}(\textbf{x})$ must adapt to any differences between $P_{source}$ and $P_{target}$ with access only to $x_t$.

Given a pretrained model, LeanTTA updates and adapts normalization layers to target statistics while maintaining its performance on the source domain. We discuss the four key steps below.
%: (1) extract and stabilize incoming statistics, (2) calculate sample divergence, (3) balance source and target statistics, and (4) normalize and \textit{reset} the model to adapt to the next sample without any chance of model collapse.

\paragraph{\textbf{Extract Statistical Information: }}

We extract statistical information from the input to each normalization layer \( x_t^l \). We then calculate the mean (\(\mu_t \in \mathbb{R}^{c}\)) and variance (\(\sigma_t^2 \in \mathbb{R}^{c}\)): $\mu_t \leftarrow \frac{1}{H \times W} \sum x_t^l$, 
$\sigma_t^2 \leftarrow Var(x_t^l)$. $H$ and $W$ represent the height and width of the input data per-channel, respectively, where $c$ is the number of channels. The calculated target statistics are combined with the saved source statistics at that layer $(\mu_s, \sigma_s^2)$ using a weighted average, parameterized by $\tau$: $\mu_b \leftarrow \tau \mu_s + (1 - \tau) \mu_t$, $\sigma_b^2 \leftarrow \tau \sigma_s^2 + (1 - \tau) \sigma_t^2$.

Target statistics, though informative, are highly variable between single data samples and cannot directly replace source statistics. Instead, we use an estimated moving average via $\tau$, which determines the weight assigned to the source ($\tau$) and the target statistics ($1-\tau$). 

% Across datasets and models, we found that $\tau=0.9$ (simulating momentum over 9 source samples and 1 test-time sample) strikes a reasonable balance: target statistics have sufficient influence to improve accuracy, while source statistics mitigate possible accuracy losses from noise or outliers. Other selections could yield higher accuracy on different shift severities, but require foresight into the nature of the shift.

\paragraph{\textbf{Calculate Sample Divergence: }}
To effectively adapt to single data points, we calculate a distance measure, $d$, incorporating Mahalanobis distance~\citep{de2000mahalanobis} to measure the divergence between source and our now-stabilized target distributions:
% \citep{podolskiy2021revisiting,colombo2022beyond}
% \vspace{em}
$\label{eqn:mahalanobis}
d \leftarrow 1 - e^{-(\mu_b - \mu_s)^T(\Sigma_s^2)^{-1}(\mu_b - \mu_s)}$. Composing the $\Sigma_s^2$ as a diagonal variance matrix $\in \mathbb{R}^{c\times c}$, inversion can be replaced by faster simple division. We apply Mahalanobis distance as it accounts for the variance structure of both source and target distributions, and has been applied in  SOTA out-of-distribution detection tasks.

\paragraph{\textbf{Balance Source and Target: }}
The Mahalanobis distance indicates distribution shift severity, informing whether to give more weight to the source or stabilized incoming statistics. For severe shifts, more emphasis is placed on source statistics; less-severe shifts place more emphasis on stabilized incoming statistics: $\mu_{\text{new}} \leftarrow d\lambda\mu_s + (1 - d\lambda)\mu_b$, 
$\sigma_{\text{new}}^2 \leftarrow d\lambda\sigma_s^2 + (1 - d\lambda)\sigma_b^2$.

$\lambda$ ensures that even when the Mahalanobis distance $d$ is 1, some of the target statistic is still incorporated to improve accuracy and ensure a smooth transition between source and target statistics. (See Appendix for justification of hyperparameter selection).

\paragraph{\textbf{Normalize and Reset}:}

The final step involves normalizing the input data $x_t^l$ using the updated statistics, resulting in the output $x_t^{l+1}$: $x_t^{l+1} = \gamma \left(\frac{x_t^l - \mu_{\text{new}}}{\sigma_{\text{new}}}\right) + \beta.$ We then reset the model per-sample, preventing catastrophic forgetting and ensuring robustness against rapid data shifts.

% \vspace{-2em}
\paragraph{\textbf{Partial Fusion Strategy:}}\label{subsec:partial_fusion}

Quantized on-device backpropagation is impractical in standard deployments, requiring specialized kernels~\citep{lin24_tt,jia2024tinytta}. The simplicity of our method makes quantization straightforward---in particular, layer fusion, which is commonly incorporated into compression pipelines. We also employ \textbf{partial layerwise adaptation} alongside \textbf{layer fusion}. Fusing convolutional and normalization layers reduces both memory cost and compute time, but it obscures batch statistics necessary for TTA. Our proposed partial-adaptation strategy fuses only deeper layers, then updates only unfused layers, allowing accuracy gains to coexist with fusion. Given that shallow layers tend to learn low-level, corruption-related features as opposed to the more structural features learned in deeper layers~\citep{huang17_instyle}, this is a theoretically sensible method.

\section{Results}
\subsection{Baseline  Comparisons}
We establish the performance of existing TTA baselines under both an abruptly-changing, data-scarce regime, where only \textbf{one data sample per-distribution-shift} exists, and a more common gradual-shift regime, where the corruption changes gradually. Three datasets with domain shift were used: image-based datasets CIFAR10-C, ImageNet1k-C~\citep{hendrycks19_dsets}, and the audio datasets BirdVox-DCASE-20k (train) + Warblr (test)~\citep{stowell18_birds}. The Warblr dataset contains citizen-recorded birdsong with natural abrupt shifts; we shuffled the image dataset to achieve the same effect. For Warblr, we report F1 score due to class imbalance. ResNet18~\citep{he16_resnet} and VGGish~\citep{vggish} models were tested for image and audio, respectively. For timing and memory, we deployed all baselines on a Raspberry Pi Zero 2W.

Beginning with \textbf{abrupt shifts} (Table~\ref{tab:resnet_performance}), we demonstrate that, even at batch size 64, backpropagation may fail to achieve accuracies comparable to adaptation-free models, illustrating how entropy loss suffers when test-time statistics are inconsistent (abrupt domain shift) or noisy (small batches).

Under abrupt shifts, our method achieves \textbf{the only consistent accuracy improvements} (11.8\% and 4.6\% error reduction for CIFAR10-C and ImageNet-C). At the same time, our backpropagation-free method allows us to easily quantize and deploy LeanTTA on CPU-only devices. Post-quantization, our model requires 11.2 MB, an order-of-magnitude less than batch-64 Tent and EATA.

Though MECTA at batch-16 (in lieu of 64 per author recommendation) runs faster than LeanTTA with batch-1, high timing variance exposes an underlying problem: with noisy gradients, many batches are deemed detrimental for adaptation and skipped by MECTA, limiting adaptation efficacy. 

TTA is rarely evaluated over binary classification tasks such as the Warblr dataset. Here, Tent outperforms LeanTTA with batch-64; similarly to trends CIFAR10 and ImageNet, performance on Warblr is good when batch sizes are large. However, we found backpropagation with batch size 64 on the Raspberry Pi was impractical due to runtime and memory constraints.

The relatively improved performance of baselines under \textbf{gradual shift} at batch size 64 (Table~\ref{tab:resnet_performance}) demonstrates the reliance of most TTA algorithms on statistically even batches. While LeanTTA may not be the best-performing algorithm under this paradigm, it consistently demonstrates accuracy improvements at a low memory footprint and batch size of one. Our method is well-suited for consistent, resource-efficient and fast adaptation but may not match traditional TTA when full resources are available or when statistics are less noisy; engineers must select the best choice for their hardware and use case.

Overall, we demonstrate that \textit{existing baselines underperform in both accuracy and system performance under the realistic deployment scenarios of data scarcity and edge device constraints}. LeanTTA avoids this trap by utilizing stabilized forward-only statistics updates to \textbf{consistently} improve accuracy under system constraints and quantization.

\begin{table}[!ht]
    \centering
    \caption{LeanTTA versus baselines on different batch sizes (BS). \new{N=5 trials.} CIFAR10 and ImageNet report accuracy under abrupt and gradual shifts on ResNet18; Warblr reports scaled F1 on VGGish. Memory (MB) and Runtime (s) are reported for ResNet18. \textcolor{gray}{Grey = accuracy decrease}, \textbf{Bold = best}, \underline{Underline = runner-up}. We did not run MECTA on Warblr due to MECTA's implementational incompatibility with a binary classifier.}
    \resizebox{\textwidth}{!}{
    \begin{tabular}{llcc cc c cc}
    \toprule
     & & \multicolumn{2}{c}{CIFAR10} & \multicolumn{2}{c}{ImageNet} & Warblr & \multicolumn{2}{c}{Efficiency} \\
    \cmidrule(lr){3-4} \cmidrule(lr){5-6} \cmidrule(lr){8-9}
     & BS & Abrupt & Gradual & Abrupt & Gradual & (f1) & Mem (MB) & Time (s) \\
    \midrule
    None (32) & 1 & 76.3$\pm$0.0 & 76.3$\pm$0.0 & 34.1$\pm$0.5 & 34.5$\pm$0.4 & 71.1$\pm$0.0 & 44.7 & 0.11$\pm$0.04 \\
    None (8) & 1 & 75.9$\pm$0.1 & 75.9$\pm$0.1 & -- & -- & 70.1$\pm$0.0 & 11.2 & 0.16$\pm$0.0 \\
    \midrule
    \multirow{2}{*}{Tent} 
     & 1 & \textcolor{gray}{9.8$\pm$1.6} & \textcolor{gray}{9.0$\pm$0.0002} & \textcolor{gray}{0.12$\pm$0.02} & \textcolor{gray}{0.47$\pm$0.04} & \textcolor{gray}{47.1$\pm$26.1} & 48.0 & 0.32$\pm$0.19 \\
     & 64 & \textcolor{gray}{69.6$\pm$1.7} & 78.4$\pm$3.73  & \textcolor{gray}{32.3$\pm$0.5} & 45.4$\pm$0.4 & \textbf{73.2$\pm$0.16} & 101.1 & 0.83$\pm$0.20 \\
    \midrule
    \multirow{2}{*}{EATA} 
     & 1 & \textcolor{gray}{9.0$\pm$0.0} & \textcolor{gray}{8.96$\pm$0.0}  & \textcolor{gray}{0.10$\pm$0.05} & \textcolor{gray}{0.05$\pm$0.01} & \textcolor{gray}{44.1$\pm$3.7} & 47.2 & \underline{0.15$\pm$0.22} \\
     & 64 & \textcolor{gray}{75.3$\pm$0.74} & \textbf{83.0$\pm$1.1} & \underline{37.0$\pm$0.13} & \textbf{46.7$\pm$0.4} & \textcolor{gray}{69.1$\pm$0.0} & 100.3 & 0.44$\pm$0.18 \\
    \midrule
    \multirow{2}{*}{DeYO} 
     & 1 & \textcolor{gray}{9.96$\pm$0.2} & \textcolor{gray}{10.1$\pm$0.1} & \textcolor{gray}{0.11$\pm$0.02} & \textcolor{gray}{0.13$\pm$0.02} & \textcolor{gray}{60.5$\pm$0.0} & 47.2 & 0.13$\pm$0.18 \\
     & 64 & \textcolor{gray}{70.2$\pm$1.2} & 76.6$\pm$0.6 & 35.7$\pm$0.2 & 45.0$\pm$0.5 & \textcolor{gray}{69.0$\pm$0.1} & 100.3 & 1.21$\pm$0.09 \\
    \midrule
    \multirow{2}{*}{MECTA} 
     & 1 & \textcolor{gray}{72.4$\pm$0.4} & 78.5$\pm$0.3 & \textcolor{gray}{15.3$\pm$0.8} & \textcolor{gray}{17.1$\pm$0.2} & -- & 46.4 & 0.21$\pm$0.06 \\
     & 16 & \textcolor{gray}{75.4$\pm$0.3} & \underline{80.5$\pm$0.3} & \textcolor{gray}{32.2$\pm$0.7} & 41.0$\pm$0.5 & \textcolor{gray}{69.4$\pm$0.08} & 59.8 & \textbf{0.10$\pm$0.14} \\
    \midrule
    \multirow{2}{*}{SAR} 
     & 1 & \textcolor{gray}{10.02$\pm$0.2} & \textcolor{gray}{10.2$\pm$0.1} & \textcolor{gray}{0.12$\pm$0.02} & \textcolor{gray}{0.12$\pm$0.03} & \textcolor{gray}{57.8$\pm$1.2} & 47.2 & 0.62$\pm$0.11 \\
     & 64 & \textcolor{gray}{76.2$\pm$0.6} & 79.8$\pm$0.5 & 35.7$\pm$0.4 & \underline{45.9$\pm$0.43} & \textcolor{gray}{69.2$\pm$0.1} & 100.3 & 2.00$\pm$0.16 \\
    \midrule
    TEMA & 1 & \textcolor{gray}{63.4$\pm$0.1} & 78.1$\pm$1.8 & \textcolor{gray}{28.9$\pm$0.2} & 43.1$\pm$0.6 & \textcolor{gray}{66.2$\pm$0.1} & \underline{44.7} & 3.89$\pm$0.11 \\
    \midrule
    Ours (32) & 1 & \textbf{79.1$\pm$0.0} & 79.1$\pm$0.0 & \textbf{37.1$\pm$0.5} & 37.1$\pm$0.7 & \underline{72.5$\pm$0.0} & \underline{44.7} & 0.21$\pm$0.01 \\
    Ours (8) & 1 & \underline{78.8$\pm$0.0} & 78.8$\pm$0.0 & -- & -- & 71.7$\pm$0.0 & \textbf{11.2} & 0.19$\pm$0.02 \\
    \bottomrule
    \end{tabular}
    } % resize box    
    \label{tab:resnet_performance}
\end{table}

\subsection{Layer-wise Updates with Adaptive Fusion}

Building on the advantages of our stateless, forward-only method, we found that we can achieve further computational gains while preserving (if not improving) accuracy, by applying LeanTTA to a partial subset of normalization layers within the model, allowing for the adaptive fusion of the unadapted layers. Fusing the deep and adapting the shallow halves of ResNet18 nets a 1.6x speedup while effectively matching the accuracy gains of full adaptation (Table~\ref{tab:fusion}).

\begin{table}[]
\centering
\caption{Accuracies and timings for ResNet18 with varying levels of fusion and adaptation.}
\label{tab:fusion}
\resizebox{0.6\textwidth}{!}{
\begin{tabular}{lcccc}
    \toprule
    & CIFAR10      & CIFAR100     & ImageNet     & Timing (s)              \\
    \midrule
    No Fusion  & 78.9$\pm$1.2 & 50.4$\pm$3.8 & 37.4$\pm$0.4 & 0.184$\pm$0.016   \\
    Half-Fused & 78.7$\pm$0.2 & 50.7$\pm$4.3 & 36.1$\pm$0.2 & 0.118$\pm$0.020   \\
    All-Fused  & 75.7$\pm$0.2 & 48.7$\pm$4.1 & 34.3$\pm$0.2 & 0.065$\pm$0.008 \\
    \bottomrule
\end{tabular}
}
\end{table} 

Our novel adaptation-plus-fusion method further improves the resource efficiency of LeanTTA while retaining our previously proven consistent accuracy gains. We also reveal an interesting trait of adaptation --- effectiveness appears to saturate in the shallow half of the vision models (Table ~\ref{tab:fusion}). Further investigation of the differential effectiveness of TTA on different layers is needed.

\section{Discussion and Conclusion}
In this work, we first identify gaps in the suitability of existing TTA methods for efficient edge deployment, namely, high memory and data consistency requirements and reliance on backpropagation, which may not be accurate under quantization. We then propose LeanTTA, a stateless, forward-only instance-based adaptation method combined with partial layerwise adaptation which matches or beats existing baselines under realistic abrupt data shifts while maintaining a minimal memory footprint and low runtimes.

% \begin{ack}
% Use unnumbered first level headings for the acknowledgments. All acknowledgments

% Do {\bf not} include this section in the anonymized submission, only in the final paper. You can use the \texttt{ack} environment provided in the style file to automatically hide this section in the anonymized submission.
% \end{ack}

\bibliographystyle{plainnat}
\bibliography{bib}

\newpage
\appendix

\section{Appendix}

\subsection{Extended Related Work}
\label{app:related}

\paragraph{TTA Methods.}
A number of diverse approaches to TTA have been proposed in literature, from human-in-the-loop reinforcement learning to optimize normalization statistics~\citep{lee25_bitta}, to adaptation of the sensor settings based on model outputs~\citep{baek25_camset} (refer to~\citet{liang25_litrev} for a detailed survey). Here, we focus on methods which modify model parameters at test-time in a traditional model deployment setting. The restrictive data assumptions inherent in straightforward entropy minimization undermine its robustness in practical applications~\citep{wang21_tent}. Prior research efforts address several of these limitations. CoTTA~\citep{wang22_cotta} tackles catastrophic forgetting and continuous shifts, while NOTE~\citep{gong23_note} and RoTTA~\citep{yuan23_rotta} both reserve batches of data to simulate independent and identically distributed (i.i.d.) data even under non-i.i.d. conditions. \new{MEMO~\citep{zhang22_memo} adapts to a single test point by minimizing the marginal entropy over augmented copies of it.} While these methods improve Tent's robustness, they rely on entropy-loss backpropagation, meaning their accuracy improvements depend on the optimizer and hyperparameters used~\citep{zhao23_pitfalls}; using entropy loss could also induce model collapse~\citep{niu24_foa}.
Being backpropagation-dependent means that they are, like Tent, much more memory-intensive than standard inference.
\new{Evaluation setups matter as well: \citet{alfarra24_time} show that, once adaptation speed is accounted for, simple and fast methods can outperform more sophisticated but slower ones; Tempora~\citep{sreeram2026tempora} shows that method rankings from conventional evaluation change under latency constraints; BoTTA~\citep{danilowski25_botta} benchmarks TTA under on-device data and resource constraints; and MORPHEUS~\citep{danilowski26_morpheus} observes that no single method wins across shifts.}

\paragraph{Resource-efficient TTA.}
Some resource-efficient TTA methods attempt to reduce the depth or frequency of backpropagation while also controlling for the noisiness of entropy as a minimization target. EATA~\citep{niu22_eata} and SAR~\citep{niu23_sar} backpropagate selectively to reduce computation of unhelpful weight updates and avoid noisy gradients. Extending these works, SOTTA~\citep{gong23_sotta} filters out noisy samples and accuracy-damaging gradients, while DeYO~\citep{lee24_deyo} adds an additional filtering metric which uses structural perturbation to boost detection of distribution shift.

MECTA~\citep{hong23_mecta} takes a different approach by pruning the gradient cache of batch normalization layers prior to backpropagation; SURGEON~\citep{ma25_surgeon} proposes an architecture-agnostic activation pruning method to reduce memory consumption during backpropagation. \new{EcoTTA~\citep{song23_ecotta} attaches lightweight meta networks to a frozen backbone to shrink the activations stored for backpropagation.} AdaShadow~\citep{fang2024adashadow} reduces latency by separately predicting the latency and importance of model layers in adaptation, then optimizing for the best schedule of layers to activate for adaptation. However, these methods require a sufficiently large batch of samples and/or multiple passes through models; in addition, all aforementioned methods rely on backpropagation, which means their worst-case computational cost is still on-par with Tent and may involve complex implementations if quantized and moved on-device~\citep{lin24_tt,jia2024tinytta,kwon2023lifelearner,kwon2024tinytrain}.

\new{For edge devices, where models are already deployed for inference on microcontrollers~\citep{kwon_yono_ipsn22,jia2023ur2m,JIA2026ur2mjournal}, TinyTTA~\citep{jia2024tinytta} adapts early-exit ensembles in a batch-agnostic strategy. DBU~\citep{xiao2026efficient} decouples the update of BN statistics from that of BN affine parameters so that the memory-heavy updates run at small batch sizes, and EmbodiTTA~\citep{ma26_emboditta} adapts only when a shift is detected. In addition, quantization and compression add further complication. Specifically, TTAQ~\citep{xiao24_ttaq} addresses the performance degradation of post-training quantization under continual domain shift, and \citet{corti2026interaction} show that the gap between TTA and supervised fine-tuning widens as structured compression increases.}

\paragraph{Backpropagation-free TTA.}
Inference-only adaptation methods comprise a divergent branch of TTA methods. Being backpropagation-free, their memory and energy consumption yield the same order of magnitude as inference.

\citet{li16_adabn} propose directly updating the frozen source statistics in normalization layers of a model with statistics calculated over the entirety of the target data. A more practical approach utilizes only mini-batch target statistics~\citep{nado20_predbn,benz21_rstats}, though this method fails as batch sizes approach one. \citet{schneider20_navg} extend the original approach by reducing the target data required through the use of a weighted average of source and target statistics. Further reducing the target data needed down to a single image, SITA~\citep{khurana21_sita} and InTEnt~\citep{dong24_intent}, respectively, use feature augmentation and entropy-weighted integration across a hyperparameter space, though both methods require multiple passes through the entire model per-adaptation. \new{DUA~\citep{mirza22_dua} also adapts normalization statistics without backpropagation, but accumulates them over the test stream, so its estimate depends on previously seen samples.} FOA~\citep{niu24_foa} enables forward-only adaptation but is ViT-specific, relies on a class token, and still requires large batch sizes or cached historical data. \new{ZOA~\citep{deng25_zoa} reports that quantized models are more sensitive to domain shift than their full-precision counterparts, and adapts quantized CNNs and ViTs through zeroth-order optimization with two forward passes per sample; FOZO~\citep{wang26_fozo} and EVA-0~\citep{chen26_eva0} likewise rely on zeroth-order, forward-only optimization.} TEMA~\citep{su24_tema} takes a different approach that involves two forward passes, trading off the hyperparameter search for a long warm-up period of roughly 1,000 test images for batch size 1.

\new{Other methods act on classifier outputs or embeddings: T3A~\citep{iwasawa21_t3a} adjusts the classifier with prototypes built from past test samples, LAME~\citep{boudiaf22_lame} corrects the model's outputs rather than its parameters, NEO~\citep{murphy26_neo} re-centers target embeddings at the origin, and PEA~\citep{ma26_pea} aligns intermediate embeddings with two forward passes, using an exponential moving average of test-time statistics.}

\paragraph{Novelty of LeanTTA.}
Our proposed method sits at the intersection of consistent accuracy improvement, low memory footprint, and fast runtime, giving it several advantages over existing baselines. Like other methods, LeanTTA is backpropagation-free, but it eschews ensembles or augmentations and adapts dynamically per data point — even if points are sampled from vastly different domains — independent of batch size or previous data. Our statistics stabilization strategy eliminates the need for a long warm-up period, and our per-data-point reset prevents any chance of long-term model collapse. Finally, LeanTTA improves both accuracy and computational efficiency by proposing adaptation on a subset of the model and fusing the non-adaptive layers.

\subsection{Methods}

\begin{figure}[t]
    \centering
    \includegraphics[width=\linewidth]{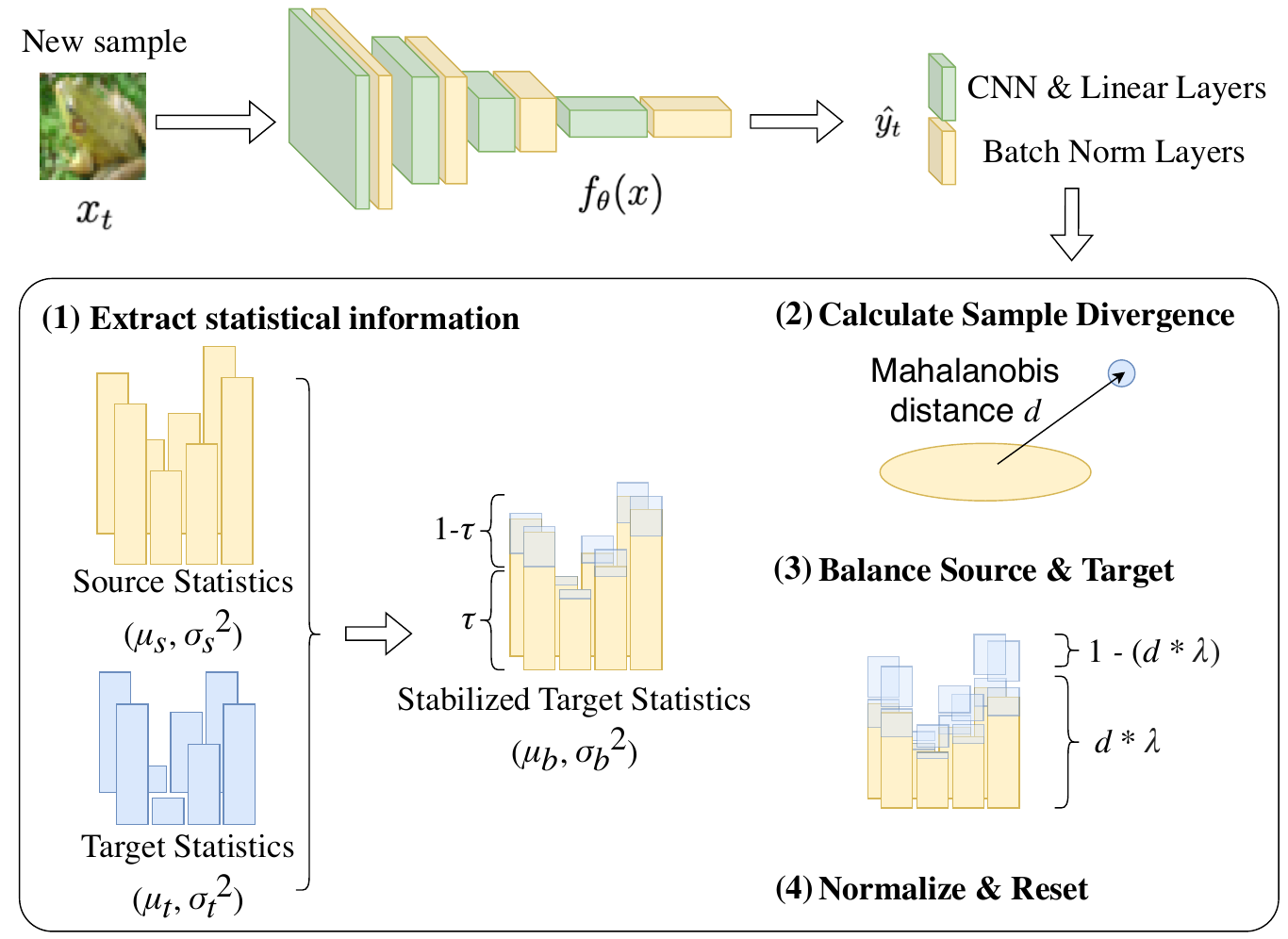}
    \caption{Proposed updates to the normalization layer. For each intermediate activation, target statistics are (2) recorded and stabilized using source statistics. The (2) Mahalanobis distance, $d$, estimates how far out-of-distribution the stabilized statistics have moved at each layer. The source and target statistics are then (3) balanced and recombined according to the scaled distance and used for (4) normalization.}
    % \chj{why pi why lambda...replotting it ing}/chj{fixed}
    \label{fig:method}
\end{figure}

We illustrate the methodology behind LeanTTA visually in Figure~\ref{fig:method}.

As an algorithm, LeanTTA is designed to be forward-only and quantization-compatible, making it well-suited for resource-limited devices. It requires \textbf{only a single forward pass with a batch size of one} to improve accuracy. As each data instance progresses through the model's normalization layers, statistics are updated sequentially. Once the intermediate representation is computed at a layer, the statistics are \textbf{immediately reset}, preparing the system for the next data point and completely mitigating any concerns around catastrophic forgetting — the model returns to its original state at each iteration. 

\begin{figure*}[t]
    \centering
    \includegraphics[width=1\linewidth]{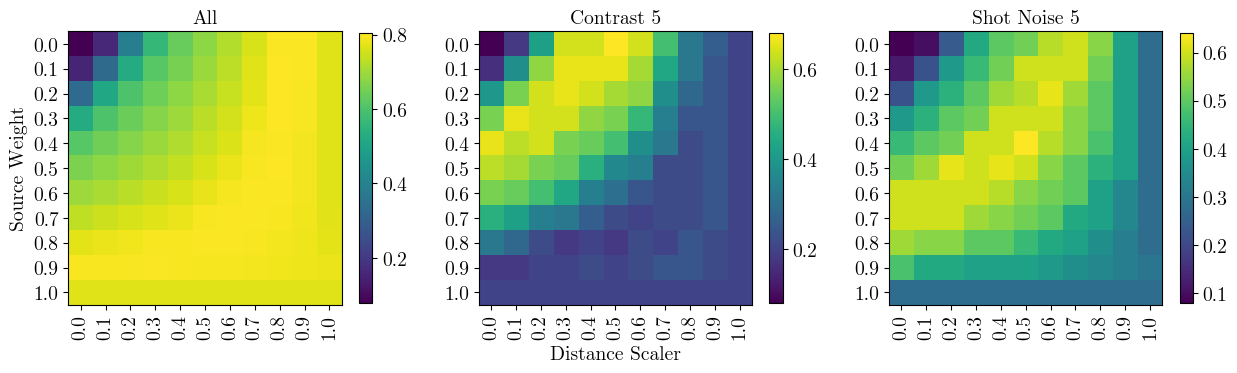}
    \caption{Hyperparameter space search for Source Weight $\tau$ and Distance Scaler $\lambda$  with ResNet18 applied to CIFAR10-C. From left to right, ``All'' shows accuracies across hyperparameter choices from all 19 categories and 5 severity levels, whereas ``Contrast 5'' and ``Shot Noise 5'' show accuracies from the corresponding noise type and severity. Lighter shaded squares indicate higher accuracy.}
    \label{fig:hyperparams}
\end{figure*}

Across datasets and models, we found that $\tau=0.9$ (simulating momentum over 9 source samples and 1 test-time sample) strikes a reasonable balance: target statistics have sufficient influence to improve accuracy, while source statistics mitigate possible accuracy losses from noise or outliers (Figure ~\ref{fig:hyperparams}). Other selections could yield higher accuracy on different shift severities, but require foresight into the nature of the shift.

\begin{figure}[t]
    \centering
    \includegraphics[width=\linewidth]{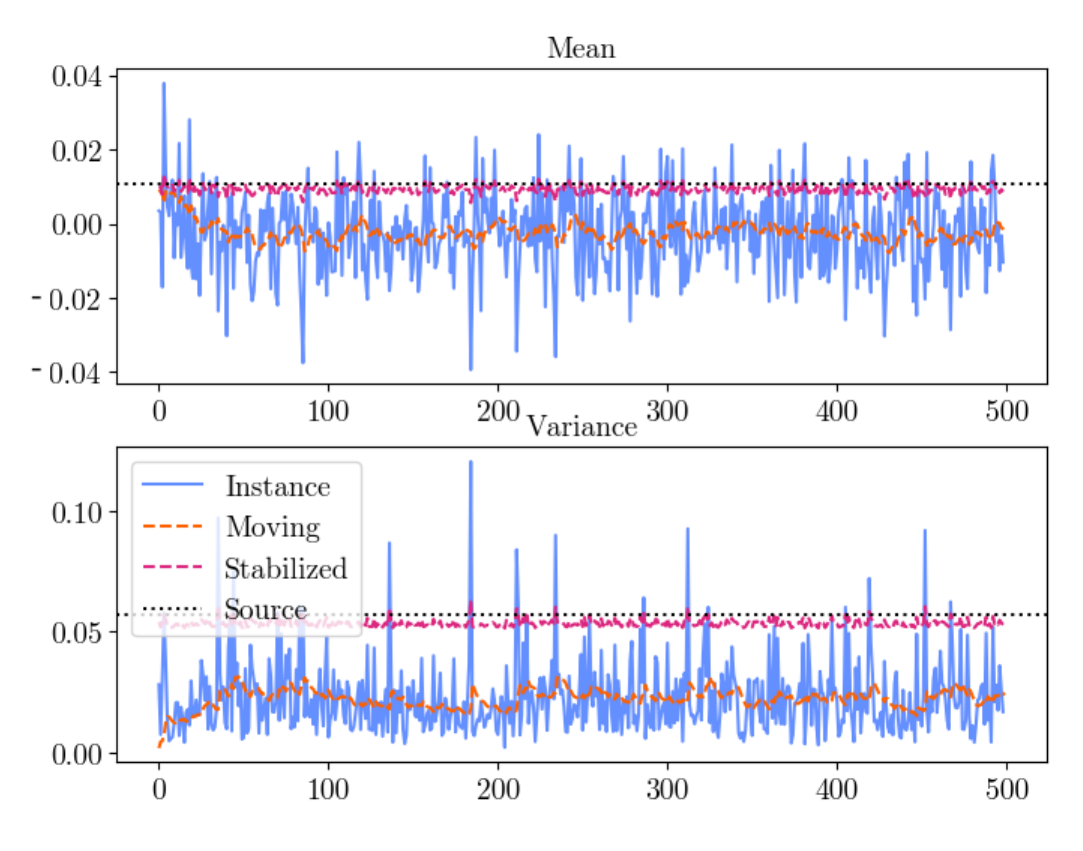}
    \vspace{-1.5em}
    \caption{Mean (top) and variance (bottom) recorded from a single layer of ResNet18 on the shift-free CIFAR10 dataset for a running average with momentum=0.9 (orange), with statistics calculated at each image (blue) and with stabilized statistics (red).}
    \label{fig:run_v_inst}
    \vspace{-1em}
\end{figure}

\subsection{Further Experiment Information}

\textbf{Model Training Details.}
For the setting of on-device computation, we employ compact architectures, namely, MobileNetV2~\citep{sandler18_mbnet} and ResNet18~\citep{he16_resnet}; we also show efficacy on a hybrid architecture through MobileViT~\citep{mehta22_mobilevit}. We utilized models pre-trained on ImageNet1k, then fine-tuned the models on the training sets of CIFAR10 and CIFAR100. For the audio modality, we use VGGish~\citep{vggish}, trained for 7,000 iterations on 16,000 BirdVox samples, with 4,000 samples held out.

\textbf{Abrupt Domain Shift Simulation.}
Non-continuous distribution shifts change abruptly, from the perspective of the model. To simulate real-world scenarios where data comes from different domains — i.e. photographs or intermittent recordings — we randomly shuffled each image dataset across shift types and severity levels, following prior works~\citep{niu23_sar,su24_tema}. For the image classification datasets, we sampled 100 images from each of the 19 corruption categories and 5 corruption levels, then shuffled those images together. The bird datasets are, by way of being from many different people, devices, and locations, \textit{naturally abruptly shifting}.

\textbf{Baselines.} We compare LeanTTA with seven SOTA TTA benchmarks: Tent~\citep{wang21_tent},  EATA~\citep{niu22_eata}, TEMA~\citep{su24_tema}, CoTTA~\citep{wang22_cotta}, SAR~\citep{niu23_sar}, DeYO~\citep{lee24_deyo}, and MECTA~\citep{hong23_mecta}, where we selected EATA as the baseline method for MECTA. We followed suggestions from their respective papers for implementation. For MECTA, we followed the advice of the authors and limited ourselves to a batch size of 16 and below for best accuracy. We applied documentation-suggested hyperparameter settings, and broadly found that single-instance batches collapsed regardless of hyperparameter settings.

\subsection{Evaluation Over Multiple Architectures}

To demonstrate the generalizability of our algorithm to multiple architectures beyond ResNet18, we also tested LeanTTA on MobileNetV2~\citep{sandler18_mbnet}, VGGish~\citep{vggish}, and MobileViT~\citep{mehta22_mobilevit}. While all utilize convolutional layers, MobileViT notably also incorporates transformer layers.

Table~\ref{tab:system} displays the memory and runtime performance across ResNet18, MobileNetV2, and VGGish on the Raspberry Pi Zero 2W. Notably, running large batch size TTA methods crashed the Raspberry Pi with VGGish and MobileNetV2. TTA over large batch sizes also demonstrated significant latency --- time \textit{per-frame}, not per-batch, stretched well into the seconds, if not tens of seconds for the more computationally intensive backpropagation-based methods. TEMA was also slow despite being forward-only, owing to its per-batch recalculation of full data distributions.

\begin{table*}[t]
  \centering
  \caption{System measurements for memory (MB) and time (seconds-per-instance) for the three evaluated models on the Raspberry Pi Zero 2W. ``-'' indicates that the Pi was unable to run adaptation due to resource constraints. ** indicates that the loss became NaN. Bold text indicates best performance, \underline{underlined} indicates second-best (with the exception of the adaptation-free baseline).}
  \label{tab:system}
  \resizebox{\textwidth}{!}{
  \begin{tabular}{llcccccc}
    \toprule
    \multirow{2}{*}{\textbf{Method}} & \multirow{2}{*}{\textbf{Variant}} & \multicolumn{2}{c}{\textbf{VGGish}} & \multicolumn{2}{c}{\textbf{ResNet18}} & \multicolumn{2}{c}{\textbf{MobileNetV2}} \\
    \cmidrule(lr){3-4} \cmidrule(lr){5-6} \cmidrule(lr){7-8}
    & & Memory & Time & Memory & Time & Memory & Time \\
    \midrule
    \multirow{2}{*}{None} & BS 1 & 18.7 & 1.46$\pm$1.40 & 44.7 & 0.113$\pm$0.04 & 8.95 & 0.18$\pm$0.11 \\
    & int8 & 4.77 & 0.48$\pm$0.10 & 11.2 & 0.159$\pm$0.0 & 2.24 & 0.50$\pm$0.06 \\
    \midrule
    TEMA & BS 1 & 18.7 & 4.13$\pm$0.60 & 44.7 & 3.89$\pm$0.11 & 8.95 & 19.6$\pm$0.16 \\
    \midrule
    \multirow{3}{*}{Tent} & BS 1 & 78.2 & 2.59$\pm$1.49 & 48.0 & 0.319$\pm$0.19 & 36.0 & 0.45$\pm$0.30 \\
    & BS 16 & 306.8 & - & 60.7 & 0.402$\pm$0.12 & 138.4 & 4.4$\pm$0.50 \\
    & BS 64 & 1038.2 & - & 101.1 & 0.830$\pm$0.20 & 466.1 & - \\
    \midrule
    \multirow{3}{*}{EATA} & BS 1 & 63.5 & \underline{1.52$\pm$1.17} & 47.2 & 0.147$\pm$0.22 & 29.3 & \textbf{0.219$\pm$0.013} \\
    & BS 16 & 292.0 & - & 59.8 & \textbf{0.088$\pm$0.16} & 131.7 & 3.44$\pm$0.44 \\
    & BS 64 & 1023.4 & - & 100.3 & 0.442$\pm$0.18 & 459.4 & - \\
    \midrule
    \multirow{3}{*}{CoTTA} & BS 1 & 263.8 & 55.31$\pm$5.39 & 279.8 & 53.9$\pm$4.5 & 152.4 & 14.8$\pm$1.16 \\
    & BS 16 & 1098.8 & - & 340.6 & 7.31$\pm$0.39 & 652.7 & 14.78$\pm$1.16 \\
    & BS 64 & 3770.9 & - & 535.2 & 5.78$\pm$1.84 & 2253.5 & - \\
    \midrule
    \multirow{3}{*}{DeYO} & BS 1 & 63.5 & 1.74$\pm$0.3 & 47.2 & 0.13$\pm$0.18 & 29.3 & \underline{0.22$\pm$0.008} \\
    & BS 16 & 292.0 & 30.0$\pm$1.3 & 59.8 & 0.41$\pm$0.05 & 131.7 & 6.42$\pm$0.84 \\
    & BS 64 & 1023.4 & - & 100.3 & 1.21$\pm$0.09 & 459.4 & 8.86$\pm$0.85 \\
    \midrule
    \multirow{2}{*}{MECTA} & BS 1 &  50.1 & ** & 46.4 & 0.21$\pm$0.06 & 23.3 & 0.44$\pm$0.04 \\
    & BS 16 & 210.0 & 47.8$\pm$2.2 & 59.8 & \underline{0.102$\pm$0.14} & 95.0 & 2.83$\pm$0.87 \\
    \midrule
    \multirow{3}{*}{SAR} & BS 1 & 63.5 & 5.19$\pm$0.13 & 47.2 & 0.617$\pm$0.107 & 29.3 & 0.88$\pm$0.16 \\
    & BS 16 & 292.0 & - & 59.8 & 0.74$\pm$0.06 & 131.7 & 11.7$\pm$1.6 \\
    & BS 64 & 1023.4 & - & 100.3 & 2.0$\pm$0.16 & 459.4 & 16.7$\pm$1.4 \\
    \midrule
    \multirow{2}{*}{LeanTTA} & BS 1 & \underline{18.7} & 1.54$\pm$0.34 & \underline{44.7} & 0.205$\pm$0.01 & \underline{8.95} & 0.531$\pm$0.030 \\
    & int8 & \textbf{4.77} & \textbf{0.85}$\pm$\textbf{0.20} & \textbf{11.2} & 0.185$\pm$0.021 & \textbf{2.24} & 0.947$\pm$0.01 \\
    \bottomrule
  \end{tabular}
  }
\end{table*}

Table~\ref{tab:mobivit_results} shows the performance of LeanTTA on CIFAR10 against our baseline set. We demonstrate that our system works similarly well on modern architectures, such as hybrid vision transformers. We also show that our finding of comparable or even high accuracies obtained with partial adaptation of shallow layers hold true, even in models with varied architectures. 

\begin{table}[t]
  \centering
  \caption{Accuracy on abruptly changing CIFAR10-C datasets using the MobileViT-s architecture (32 BN layers) fine-tuned to the CIFAR10 dataset. Our results are shown with adaptation across varying numbers of layers. N=5 trials. \textbf{Bold}: best, \underline{underline}: 2nd best, \textcolor{gray}{grey text}: worse than baseline.}
  \label{tab:mobivit_results}
  \begin{tabular}{l c c}
    \toprule
      \textbf{Method} & \textbf{Variant} &
      \textbf{Accuracy} \\
      \midrule
      None & Batch 1 & 72.0 $\pm$ 0.4 \\
      \midrule
      TEMA & Batch 2 & \textcolor{gray}{15.5 $\pm$ 0.4} \\
     \midrule
      \multirow{2}{*}{Tent} & Batch 16 & \textcolor{gray}{69.8 $\pm$ 1.0} \\
      & Batch 64 & 75.2 $\pm$ 0.4 \\
      \midrule
      \multirow{2}{*}{EATA} & Batch 16 & \textcolor{gray}{71.1 $\pm$ 0.7} \\
      & Batch 64 & 74.2 $\pm$ 0.2 \\
      \midrule
      \multirow{2}{*}{CoTTA} & Batch 16 & \textcolor{gray}{57.7 $\pm$ 0.5}  \\
      & Batch 64 & \textcolor{gray}{60.9 $\pm$ 1.0}\\
      \midrule
     \multirow{2}{*}{DeYO} & Batch 16 & \textcolor{gray}{71.4$\pm$0.7} \\
     & Batch 64 & 74.7$\pm$0.4 \\
     \midrule
     MECTA & Batch 16 & \textcolor{gray}{15.7$\pm$0.7} \\
     \midrule
     \multirow{2}{*}{SAR} & Batch 16 & \textcolor{gray}{70.8$\pm$0.7} \\
     & Batch 64 & 74.3$\pm$0.4 \\
     \midrule
     \multirow{4}{*}{LeanTTA (ours)} & Full Adap. & 73.8 $\pm$ 0.7 \\
      & 1st 8 & \underline{75.3 $\pm$ 0.3} \\
      & 1st 16 & \textbf{75.6 $\pm$ 0.3} \\
      & 1st 24 & 74.8 $\pm$ 0.5 \\
      \bottomrule
  \end{tabular}
\end{table}

\subsection{Fusion \& Layerwise Adaptation}
In Table~\ref{tab:fusion}, we show the comparable accuracy resultant from adaptation of only the shallow layers in a model across datasets. Our analysis underscores the utility of only conducting partial fusion --- doing so allows for not only faster adaptation, but also layer fusion, as discussed in the main body of this work.

\begin{figure}[t]
    \centering
    \includegraphics[width=0.7\linewidth]{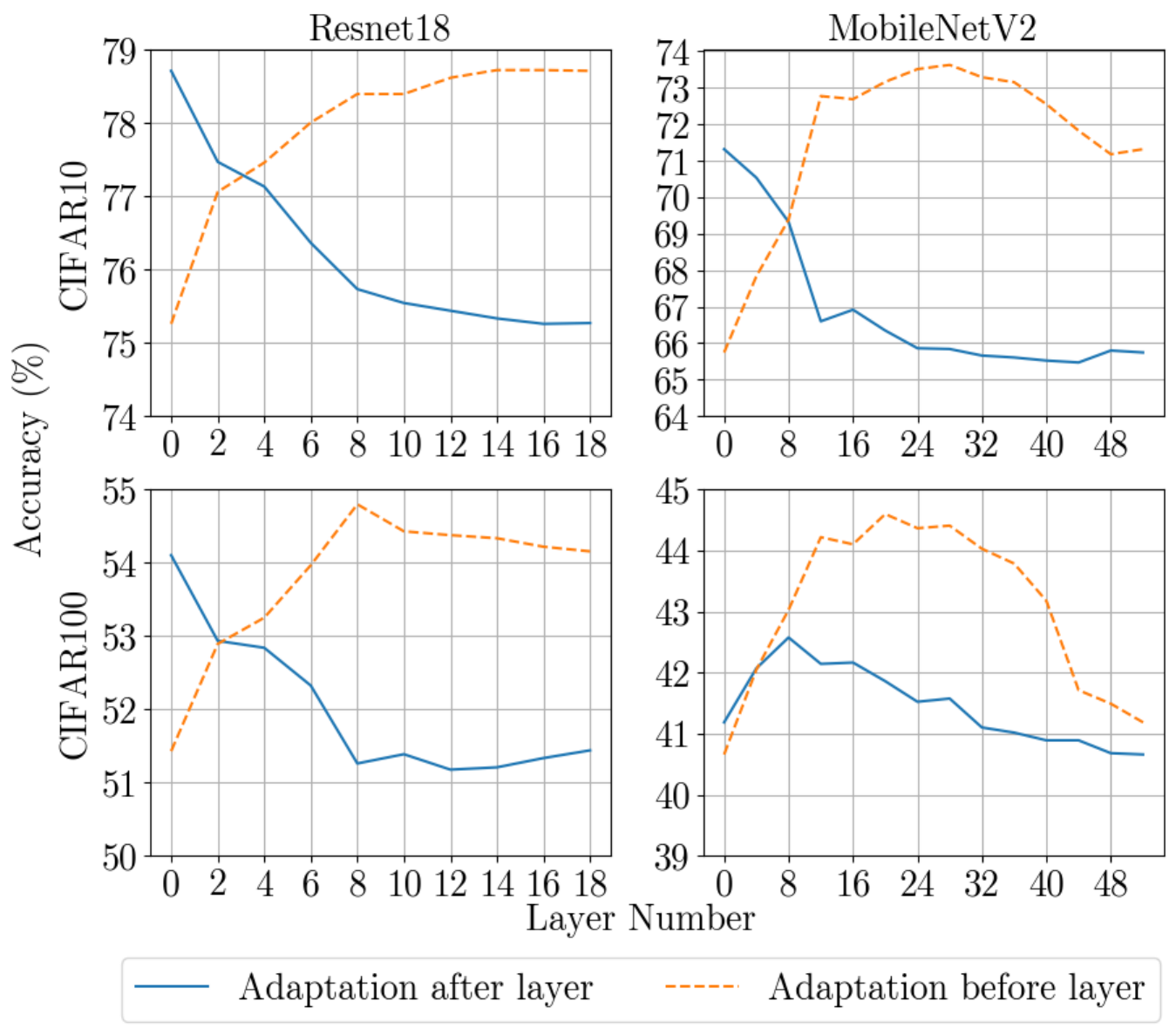}
    \caption{Layer-wise ablation analysis: (1) progressively removing adaptation from shallow layers (blue solid line) and (2) progressively adding adaptation to deeper layers (orange dotted line). Results are from full-precision (fp32) ResNet18 and MobileNetV2 evaluated on the Abrupt CIFAR10 dataset.}
    \label{fig:layer_ablation}
\end{figure}

\begin{figure}[t]
    \centering
    \includegraphics[width=0.8\linewidth]{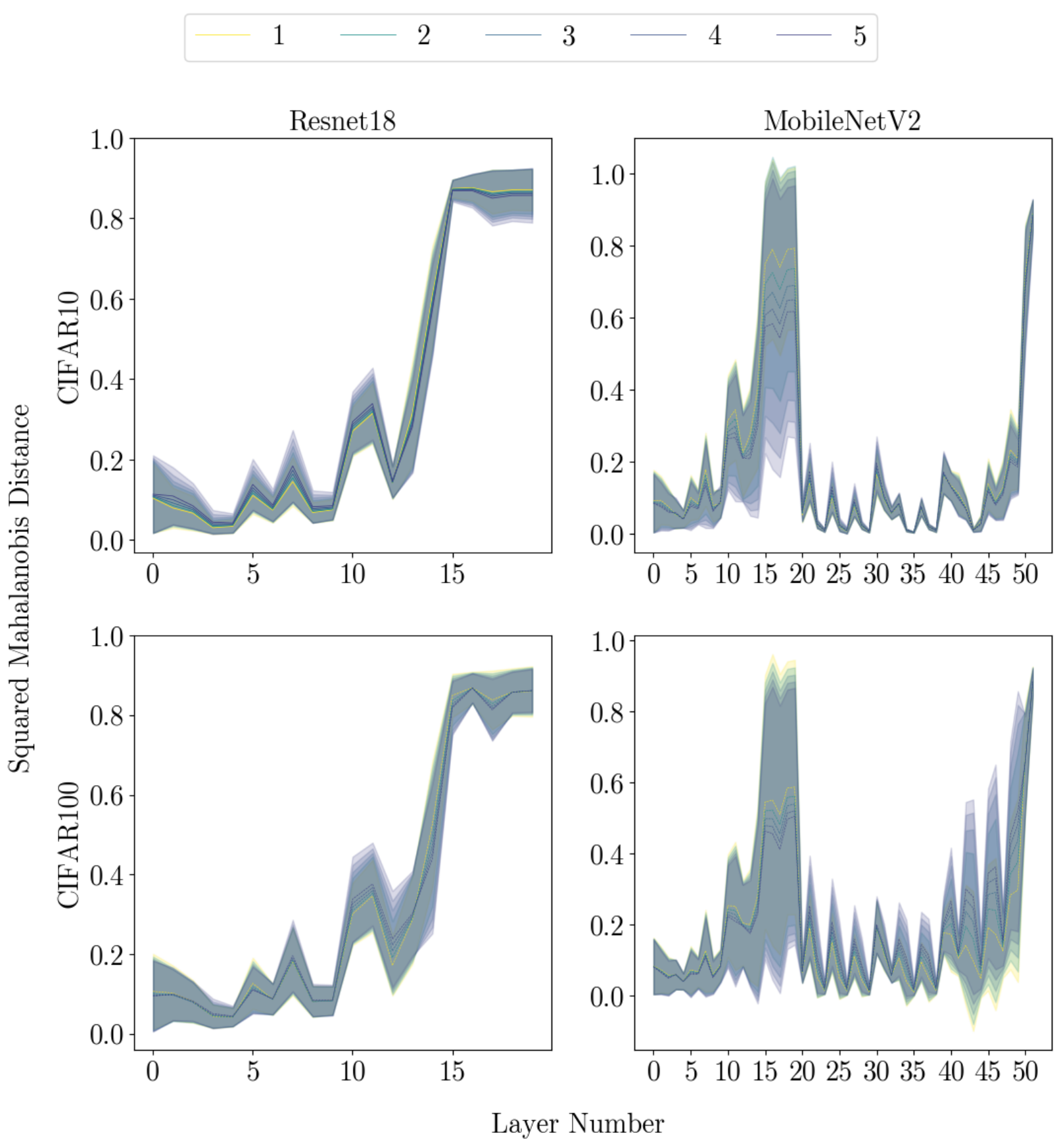}
    \caption{Layerwise Mahalanobis distance $d$ calculated after stabilization in Equation~\ref{eqn:mahalanobis}, separated by severity of distribution shift in the CIFAR10/100-C datasets (light being the least severe, dark being the most). Shading indicates one standard deviation on either side of the average over N=5 trials.}
    \label{fig:layer_dist}
\end{figure}

Skipping shallow layer adaptation reduced accuracy improvements, indicating that these layers are more important to adaptation (Figure~\ref{fig:layer_ablation}). We hypothesize that deep layers focus on larger structures, while shallow layers learn general patterns and components. Such ``stylistic'' information is related to distribution shift~\citep{benz21_rstats}. Looking at Figure~\ref{fig:layer_dist}, we see support for this hypothesis: deeper layers (for the most part) have higher divergence between train- and test-time statistics. 

\end{document}